\documentclass[procedia]{easychair}

\usepackage{doc}
\usepackage{makeidx}
\usepackage{subcaption}
\usepackage{afterpage}

\newcommand{\argmax}[1]{\underset{#1}{\operatorname{arg}\,\operatorname{max}}\;}

\def\procediaConference{BICA}

\title{Affective Facial Expression Processing via Simulation: A Probabilistic Model}

\titlerunning{Affective Facial Expression Processing}

\author{
    Jonathan Vitale\inst{1}
    \and
    Mary-Anne Williams\inst{1}
    \and
    Benjamin Johnston\inst{1}
    \and
    Giuseppe Boccignone\inst{2}
}

\institute{
  QCIS Centre - University of Technology Sydney,
  Ultimo, NSW 2007, Australia\\
  \email{jonathan.vitale@student.uts.edu.au}\\
  \email{Mary-Anne@TheMagicLab.org}\\
  \email{benjamin.johnston@uts.edu.au}\\
\and
   Dipartimento di Informatica - Universit\'a degli Studi di Milano,
   Milano, 20135 Italy\\
   \email{giuseppe.boccignone@unimi.it}\\
 }

\authorrunning{Vitale, Williams, Johnston and Boccignone}

\begin{document}

\maketitle


\begin{abstract}
  Understanding the mental state of other people is an important skill for intelligent agents and robots to operate within social environments.
  However, the mental processes involved in `mind-reading' are complex.
  One explanation of such processes is Simulation Theory --- it is supported by a large body of neuropsychological research. Yet, determining the best computational model or theory to use in simulation-style emotion detection, is far from being understood.
  In this work, we use Simulation Theory and neuroscience findings on Mirror-Neuron Systems as the basis for a novel computational model, as a way to handle affective facial expressions.
  The model is based on a probabilistic mapping of observations from multiple identities onto a single fixed identity (`internal transcoding of external stimuli'), and then onto a latent space (`phenomenological response'). Together with the proposed architecture we present some promising preliminary results.
\end{abstract}

%------------------------------------------------------------------------------
\section{Introduction}
\label{sec:introduction}
In this paper we propose a probabilistic computational theory for the \emph{detection} of emotion states based on  facial expressions. This is considered a circumscribed but important mindreading task \cite{goldman2005simulationist}, either from a neuropsychological theoretical perspective and for more application oriented areas such as social robotics and social signal processing.

From a general standpoint, `mind-reading' is the process of inferring the mental state of other people based on their facial expressions or their overt/observable behaviour. 
It plays an important role in social interaction, empathy and effective communication. 
If intelligent agents and robots have to interact with us naturally and in social situations, then it will be important for them to be able to `mind-read' people.

A large body of neuropsychological research supports Simulation Theory (ST) as a plausible account for mindreading (see \cite{goldman2005simulationist} for an in-depth review). 
According to ST, an observer arrives at a mental attribution by simulating, in his/her own mind and body, the same state as the target. 
However, while the neuropsychological account of Simulation Theory is compelling, the critical question of which computational model or theory\footnote{The ``what'' level of explanation, in the sense of Marr \cite{Marr}} might be used in simulation-style emotion detection remains poorly understood and largely unexplored \cite{goldman2005simulationist,zeng2009survey,pantic2007machine}. 
The aim of this work is to take a step to bridge this gap by building a biologically-inspired computational account of mind-reading that is suitable for artificial agents.

In particular, we seek to apply Simulation Theory to the problem of mapping an observed overt behaviour (in this work, limited to facial expressions) to a phenomenological internal latent space of the mind-reader\footnote{In this context proposed as the internal response of the subject given a particular stimulus}. In this study we will not consider the attribution of a specific mental state given such internal representation, as it is not the target of this paper.

Interestingly enough, computational approaches to affective state recognition from facial expression, focus on emotion labelling via direct inference of affect from facial expression \cite{zeng2009survey,pantic2005web}. Such approaches mostly rely on Theory-Theory (TT) based accounts. In the TT account the mindreader selects a mental state for attribution to a target based on inference from other information about the target (e.g. concerning relationships or transitions between psychological states and/or behavior of the target). The rationale for such assumption is prima facie evident: postulating theory-based mechanism only requires to link facial configurations with emotion names. 

However, this short-term computational advantage, which avoids a generative, simulation-based step, is, in our view, apparent. As Goldman and Sripada put it \cite[p. 208]{goldman2005simulationist}:  ``Simulation might be (somewhat) complex from a functional perspective, but it might be simpler from an evolutionary perspective. Simulation relies upon running the same emotional apparatus (possibly in reverse) that is already used to generate or experience the emotion. As a consequence, simulation routines do not require an organism to be outfitted with entirely new processes in order to confer an ability to recognize emotions in others". Clearly, this parsimonious stance offers some long term advantages, for instance when coming to the challenge of integrating multiple modalities of social signalling \cite{zeng2009survey}.  

The remainder of the paper is organised as follows.
In Sec. \ref{Sec:simul_account}, we provide background and motivations for a simulationist account of mind-reading. In Sec. \ref{Sec:model} we formalize a novel computational theory \cite{Marr} shaped in the form of a probabilistic model. In Sec. \ref{Sec:simul} we provide one possible computational implementation of the proposed model and some preliminary experimental results.
We conclude with a summary in Sec. \ref{Sec:concl}.

\section{Background, Motivations and Main Contributions of the Proposed Approach}
\label{Sec:simul_account}

The experience of emotion occurs when a complex state of the organism is accompanied by variable degrees of awareness, variously indicated as `appraisal'.
Two levels of emotion appraisal can be distinguished \cite{lambie2002consciousness}: a first-order phenomenological state and a conscious second-order awareness. 
Both states can be either self-directed (first-person perspective) or world-directed (third person perspective). 
The content of the first-order phenomenological state is physical and visceral, centered on one's body state and related neural underpinnings, that will be generically referred in the following discussion as $\mathbf{X}$.
By contrast, the content of second-order conscious awareness can be either propositional or non-propositional, which we will denote $\mathbf{M}$. 
In this work, we will be concerned with first-order emotion experience, thus relying on the $\mathbf{X}$ representation.
To make this intention clearer, in the rest of this paper we will use the term \emph{detection} to describe the process of deriving a first-order phenomenological state $\mathbf{X}$ given an internal representation of an observed face expression, and \emph{recognition} that of determining a second-order labeling $\mathbf{M}$.

Now, we focus on the simulationist account of emotion \emph{detection}, which can be summarised as follows:
\begin{enumerate}

\item In a certain situation, a subject (the \emph{actor}) experiences an internal state $\mathbf{X}_{act}$. The internal state may be either triggered by an external event and/or mentally induced (e.g. a particular memory).
\item The internal state elicits a corresponding behaviour $\mathbf{Y}_{act}$ (e.g. facial expression, gesture, heart beat, \dots).
\item A second subject, (the \emph{observer}) selects a mental state $\mathbf{M}_{obs}$ for attribution after enacting within himself the internal state $\mathbf{X}_{obs}$ in question by using as evidence the actor's behaviour observable by $\mathbf{Y}_{act}$ \cite{goldman2005simulationist}.
\end{enumerate}

\begin{figure}[tb]
	\begin{centering}
	\includegraphics[width=0.5\textwidth]{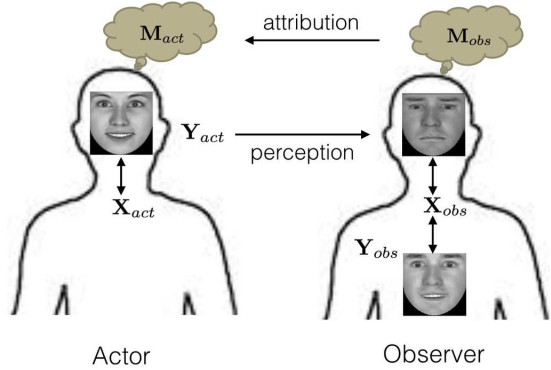}
	\caption{The problem of emotion detection from face expression}
	\label{fig:problem}
	\end{centering}
\end{figure}

\noindent This account is illustrated in Fig. \ref{fig:problem}.

There are several ways that this account might be translated into a computational theory.
Goldman and Sripada \cite{goldman2005simulationist} have devised four that have substantial plausibility and are consistent with neuropsychological evidence:
\begin{enumerate}
\item Generate-and-test models;
\item Reverse simulation models;
\item Variants of the reverse simulation model that employ an \emph{as if} loop;
\item Unmediated resonance models.
\end{enumerate}

\paragraph{Generate-and-test models}

Generate-and-test models assume that the observer starts by hypothesizing a certain internal state of the actor $\mathbf{X}_{act}$ as the possible cause of the actor's facial display $\mathbf{Y}_{act}$.
The actor proceeds to `enact' that very same state, that is, produce a facsimile of it, $\mathbf{X}_{obs}$, in his/her own system. 
If the resulting facial expression $\mathbf{Y}_{obs}$ of such simulated process matches the expression observed in the target $\mathbf{Y}_{act}$, then the hypothesized internal state is classified with a specific interpretation of a mental affective state $\mathbf{M}_{obs}$ and the observer imputes that affective state to the actor: 

\texttt{\[\mbox{Given an hypothesis } \mathbf{X}_{act} \simeq \mathbf{X}_{obs}: \mbox{ IF } \mathbf{X}_{obs} \mapsto \mathbf{Y}_{obs} \simeq \mathbf{Y}_{act} \mbox{ THEN } \mathbf{X}_{obs} \mapsto \mathbf{M}_{obs} \simeq \mathbf{M}_{act}\]}

\paragraph{Reverse simulation models}

In reverse simulation, the observer engages in the reverse direction.
Namely, given $\mathbf{Y}_{act}$, the observer activates facial muscles to imitate the actor's facial expression, $\mathbf{Y}_{obs} \simeq \mathbf{Y}_{act}$.
In order to activate such facial muscles, the observer activates its corresponding internal state $\mathbf{X}_{obs}$.
Under the assumptions that $\mathbf{X}_{obs} \simeq \mathbf{X}_{act}$, this internal experience is classified as $\mathbf{M}_{obs}$ and attributed to the actor:

\texttt{\[\mbox{Given } \mathbf{Y}_{obs} \simeq \mathbf{Y}_{act}: \mathbf{Y}_{obs} \mapsto \mathbf{X}_{obs} \mapsto \mathbf{M}_{obs} \simeq \mathbf{M}_{act}\]}

\paragraph{Reverse simulation models employing an \emph{as if} loop}

The \emph{as if} variants of reverse simulation assume that there may be direct links between a visual representation of the actor's facial expression $\mathbf{Y}_{act}$ and a somatosensory representation of ``what it would feel like'' were the observer to make that expression, i.e. $\mathbf{X}_{obs}$.
Thus, this model is similar to reverse simulation but avoids involving the facial musculature to generate $\mathbf{Y}_{obs}$:

\texttt{\[\mathbf{Y}_{act} \mapsto \mathbf{X}_{obs} \mapsto \mathbf{M}_{obs} \simeq \mathbf{M}_{act}\]}

\paragraph{Unmediated resonance models}

The rationale behind the unmediated resonance model is that observation of the actor's face directly triggers the activation of a same neural substrate $\mathbf{X}_{obs}$ in the observer associated with the internal state of the actor $\mathbf{X}_{act}$ in question. 
This is most similar to the  \emph{as if} model and there is no mediation of any kind.

At least two observations may be drawn regarding these potential computational theories. The first is that some sort of projection or cross-modal matching should be introduced to map the visual representation $\mathbf{Y}_{act}$ of actor's expression to  an internal representation of observer's own facial expression $\mathbf{Y}_{obs}$ (most likely, a proprioceptive representation, but see \cite{goldman2005simulationist} for a general discussion). This is certainly evident for the generate-and-test and reverse simulation models, but are also relevant for the \emph{as if} and unmediated resonance models (for a discuss see Sec. \ref{Sec:model}).

The second conclusion is that all models assume that the production of the relevant internal state in an observer is eventually transmitted to some cognitive center that `recognizes' or `labels' the experienced state (i.e., a $\mathbf{M}_{obs}$). This step relates to second-order emotion appraisal, which we do not consider in this paper.

In the remainder of this paper, we propose a general probabilistic framework which generalises the most salient aspects of the four approaches discussed above. Namely, we address the issue of formalising in a probabilistic latent variable space the transformations at the core of any ST approach: 
\begin{enumerate}
\item The mapping from sensed facial expression to a self-centred expression (i.e., transcoding $\mathbf{Y}_{act} \mapsto \mathbf{Y}_{obs}$);
\item The generation of a self-centred expression from a first-order phenomenological state, $\mathbf{X}_{obs}\mapsto \mathbf{Y}_{obs}$; 
\item The inverse mapping or decoding $\mathbf{Y}_{obs} \mapsto \mathbf{X}_{obs}$.
\end{enumerate} 

In literature are available some probabilistic models for motor action prediction and understanding inspired by simulation theories (see for example \cite{demiris2003distributed,dindo2011motor,wolpert2001motor}). However, to the best of our knowledge the proposed model is novel in addressing the problem of providing a computational account of simulation-style emotion detection in the context of facial expressions. In the work by Watanabe et al. \cite{watanabe2007mapping} the authors present a computational model for mapping facial expression of a caregiver to internal emotional state of a robot. Whereas the final aim of the proposed work is similarly related with this research and it shares some design ideas, it presents also some limitations that we intend to overcome with our proposed model. Among others, the generalization of the model over several different observed identities and the implementation of both a forward and inverse mechanisms able to not only detect but also generate facial expressions. In addition, this framework draws inspiration from neuroscience studies on Mirror-Neuron Systems (MNS), making it consistent with biological human findings.

\section{The Model}
\label{Sec:model}

\subsection{Rationale}

In order to model a ST process the observer has to `experience' the very same internal state of the actor \cite{goldman2005simulationist}. In this respect, all four approaches previously outlined satisfy this necessary condition, simulating the internal state using different body systems, such as specific emotion production systems, the facial musculature and somatosensory centres.

From a neuropsychological perspective,  some areas of the brain seem to be more involved in `emotional' representations of internal states (e.g. the insula or the amygdala) \cite{adolphs2002neural}, while others areas serve more as an internal action representations system (MNS), normally associated with producing actions and triggered during the observation of someone else's corresponding actions \cite{gallese2004unifying,gallese2007before}. More generally, a whole range of different mirror matching mechanisms instantiating simulation routines is likely to  be present in our brain \cite{gallese2003manifold}.

In this study, we focus on the `motor' internal representation of the perceived stimulus and we are not considering other kinds of `emotional' internal representations, more in the vein of Gallese's ``shared manifold hypothesis'' \cite{gallese2001shared,gallese2003manifold} and related findings related to the MNS \cite{gallese2004unifying,gallese2007before}. This is not much of a limitation in a simulation-style emotion detection system. In fact, as highlighted by Adolphs \cite{adolphs2002recognizing,adolphs2002neural}, the sensory-motor system appears to be critical for the recognition of emotions displayed by others, because the sensory-motor system appears to support the reconstruction of what it would feel like to be in a particular emotion, by means of simulation of the related body state \cite{adolphs2002neural}.

Another critical issue about simulation theories is the main distinction between those where actual facial movements are put into work (generate-and-test and reverse simulation) and those supported by an \emph{as if} mechanism. This distinction is related to several controversies from a neuropsychological perspective. ``Wired'' tendencies to micro-mimicking and imitation are well supported since the early work of Meltzoff and Moore \cite{meltzoff1983newborn} and Dimberg  and Thunbergs \cite{dimberg1998rapid}. In these studies, they have shown that subjects spontaneously and covertly activate facial musculature corresponding to visually presented facial expressions; meanwhile, reverse simulation is consistent with the  ``facial feedback hypothesis''  \cite{levenson1990voluntary,adelmann1989facial}.
As an alternative, facial mimicry may accompany but not actually facilitate recognition. A correlational, rather than causal, role for facial musculature in the recognition process is consistent with the results of Calder \emph{et al.} \cite{calder2000facial} and Keillor \emph{et al.} \cite{keillor2002emotional} where a patient with bilateral facial paralysis performed normally on facial based emotion recognition tasks.

Even if this should be controversial from a neuropsychological perspective, it is not a critical conceptual issue from a strict computational modeling standpoint. 
Certainly, at the ``what'' level of explanation \cite{Marr}, it is mandatory to account for the transcoding process we have previously summarized via the $\mathbf{Y}_{act} \mapsto \mathbf{Y}_{obs}$ mapping from the external visual representation of actor's expression $\mathbf{Y}_{act}$ to the internally observer-centered representation of the external stimulus $\mathbf{Y}_{obs}$ (whether it is an internal image, or a proprioceptive representation in somatosensory areas or a bare set of motor parameters).

Clearly, at the simulation/algorithmic  level (the ``how'' level in Marr's terminology \cite{Marr}) this issue can be relevant, at least for practical purposes. However, this has been a largely studied problem and elegant solutions are at hand. See for instance, in the field of robotics, Lopez and Santos-Victor's work \cite{lopes2005visual} on how to compute and learn, through self-observation, a visuo-motor map suitable to transcode visual information to motor data for hand gesture imitation tasks. To focus on the essential properties of the model, we will simply assume an internal representation $\mathbf{Y}_{obs}$ of the actor's expression in terms of an image generated by the corresponding latent space representation (see Fig. \ref{fig:problem}).

\subsection{Proposed Model}

Our probabilistic model for ST-based emotion detection is defined as follows.
Assume two interacting subjects, an actor and an observer (Fig. \ref{fig:problem}), and consider state variables  $\mathbf{Y}_{act}, \mathbf{Y}_{obs}, \mathbf{X}_{act}, \mathbf{X}_{obs},\mathbf{M}_{act}, \mathbf{M}_{obs}$ as random variables (RVs). 

Based on the experimental findings of  Gallese \cite{gallese2001shared}, these subjects are then assumed to `share' a latent manifold of first-order phenomenological states (see Sec. \ref{Sec:simul_account}). In this manifold, RV $\mathbf{X}$ takes values, and has forward and inverse mapping mechanisms $\mathbf{X}  \mapsto \mathbf{Y}$ and  $\mathbf{Y}  \mapsto \mathbf{X}$, respectively. 

We then define the following:
\begin{itemize}
  \item $P(\mathbf{Y}_{obs} \mid \mathbf{Y}_{act})$, the conditional probability density function (pdf) representing for the observer  the probability  of internally sensing $\mathbf{Y}_{obs}$ when the actor displays  $\mathbf{Y}_{act}$;
  \item $P(\mathbf{X}_{obs}| \mathbf{Y}_{obs})$, the conditional pdf representing the probability of being in an internal state $\mathbf{X}_{obs}$  given $\mathbf{Y}_{obs}$ (inverse probability);
  \item $P(\mathbf{Y}_{obs} \mid \mathbf{X}_{obs})$, the conditional pdf that the observer generates an internal facial display $\mathbf{Y}_{obs}$ given the internal state $\mathbf{X}_{obs}$ (forward probability); 
  \item $\mathcal{D}(\cdot) $, a decision or matching function comparing the ``internal" actor's display $\mathbf{Y}_{obs}$ with a simulated display $\mathbf{\check{Y}}_{obs}$.
\end{itemize}

Then, simulation-style emotion detection boils down to the following (where the symbol $\sim$ stands for the sampling operator): 

\begin{equation}
\mathbf{y}_{obs} \sim P(\mathbf{Y}_{obs} \mid \mathbf{Y}_{act}=\mathbf{y}_{act}),
\label{eq:trans}
\end{equation}

\begin{equation}
\mathbf{x}_{obs}  \sim P(\mathbf{X}_{obs} \mid \mathbf{Y}_{obs} = \mathbf{y}_{obs}),
\label{eq:code}
\end{equation}

\begin{equation}
\mathbf{\check{y}}_{obs} \sim P(\mathbf{Y}_{obs} \mid \mathbf{X}_{obs}= \mathbf{x}_{obs}),
\label{eq:gen}
\end{equation}

Eq. \ref{eq:trans} defines the transcoding process transforming an instance of  actor's face expression $\mathbf{y}_{act}$ into observer's internal representation $\mathbf{y}_{obs}$.
Eq. \ref{eq:code} defines the inverse process of experiencing/detecting state $\mathbf{x}_{obs}$ under (internal) face expression $\mathbf{y}_{obs}$.  
Eq. \ref{eq:gen} accounts for the observer's forward process of sampling his own simulated internal expression $\mathbf{\check{y}}_{obs}$ when he/she is in the latent internal state $\mathbf{x}_{obs}$. 

The function $\mathcal{D}(\cdot)$ compares the transcoded $\mathbf{y}_{obs}$ (Eq. \ref{eq:trans}) against  the internally generated expressions $\check{y}_{obs}$ (Eq. \ref{eq:gen}), and is used (via  Eq. \ref{eq:code}) to control when the matching process has converged to the most likely solution.

These three equations and $\mathcal{D}(\cdot)$ are sufficient to provide the core basis for the simulation model.

Clearly, the full picture will only be complete after attributing the actor $\mathbf{M}_{act} = m_{obs}^{\star}$, where $m_{obs}^{\star}$ is  the most likely emotion `label' instantiated by the observer through a further inferential step, by relying on the conditional pdf $P(\mathbf{M}_{obs} \mid \mathbf{X}_{obs} = \mathbf{x}_{obs}, \mathbf{C})$, where RV $\mathbf{C}$ summarizes general contextual or cultural factors. As previously discussed, the attribution issue is beyond the scope of this work.

In order to define the transcoding pdf $P(\mathbf{Y}_{obs} \mid \mathbf{Y}_{act})$  and  the forward / inverse pdfs $P(\mathbf{Y}_{obs} \mid \mathbf{X}_{obs}), P(\mathbf{X}_{obs} \mid \mathbf{Y}_{obs})$, we assume that transcoding and forward / inverse mappings  occur in two distinct latent spaces: the self-projection latent space and  the first-order phenomenological latent space.

The self-projection latent space $\mathbf{Z}$ can be conceived in terms of Bayesian latent factor regression \cite{kmurphy}:
\begin{equation}
P(\mathbf{z}) = \mathcal{N}(\mathbf{0}, \mathbf{I}_{L})
\label{eq:bfrprior}
\end{equation}

\begin{equation}
P(\mathbf{y}_{act} \mid \mathbf{z}) = \mathcal{N}(\mathbf{W}_{act}\mathbf{z} + \mu_{act}, \sigma^2_{act}\mathbf{I}_{D})
\label{eq:bfract}
\end{equation}

\begin{equation}
P(\mathbf{y}_{obs} \mid \mathbf{z}) = \mathcal{N}(\mathbf{W}_{obs}\mathbf{z} +  \mathbf{\mu}_{obs}, \sigma^2_{obs}\mathbf{I}_{D})
\label{eq:bfrobs}
\end{equation}
\noindent where $\mathcal{N}(\cdot)$ denote the Gaussian distribution; $\mathbf{W}_{act}$ and $\mathbf{W}_{obs}$ the mapping parameters for the actor and the observer, respectively; $\mathbf{\mu}, \sigma$, mean and variances;  $\mathbf{I}_{L}, \mathbf{I}_{D}$ identity matrices of dimension $L,D$ (respectively the reduced dimension of the latent space $\mathbf{Z}$ and the dimension of the vector representing the facial expression). 

\noindent Denote
 $\mathbf{W}=\begin{pmatrix}
       \mathbf{W}_{act}   \\
      \mathbf{W}_{obs}
\end{pmatrix}$,
$\mathbf{\mu}=\begin{pmatrix}
       \mathbf{\mu}_{act}   \\
      \mathbf{\mu}_{obs}
\end{pmatrix}$,
$\mathbf{\Phi}=\begin{pmatrix}
       \sigma^2_{act}\mathbf{I}_{D}  &   \mathbf{0}  \\
     \mathbf{0} & \sigma^2_{obs}\mathbf{I}_{D}
\end{pmatrix}$.
Since the model is jointly Gaussian, then $\mathbf{y}_{act}$ and $\mathbf{y}_{obs}$ are jointly Gaussian distributed, i.e. $P(\mathbf{y}_{act}, \mathbf{y}_{obs}) = \mathcal{N}( \mathbf{\mu}, \mathbf{\Sigma)}$ with mean $\mathbf{\mu}$ and covariance matrix $\mathbf{\Sigma} = \mathbf{\Phi} + \mathbf{W} \mathbf{W}^{T}$. 

In order to implement Eq. \ref{eq:trans} it is necessary the conditional distribution $P(\mathbf{Y}_{obs} =   \mathbf{y}_{obs} \mid \mathbf{Y}_{act} =   \mathbf{y}_{act})$, which again has a Gaussian distribution. Thus,

\begin{equation}
\mathbf{y}_{obs} \mid \mathbf{y}_{act}  \sim  \mathcal{N}(\widehat{\mathbf{\mu}}_{obs},  \widehat{\mathbf{\Sigma}}_{obs})
\label{eq:trans2}
\end{equation}
\newpage
\noindent where $\widehat{\mathbf{\mu}}_{obs}= \mathbf{\mu}_{obs}+ \mathbf{\Sigma}_{c}^{T}\mathbf{\Sigma}_{a}^{-1}(\mathbf{y}_{act} - \mathbf{\mu}_{act})$ and the covariance matrix 
$\widehat{\mathbf{\Sigma}}_{obs} = \mathbf{\Sigma}_{b} -  \mathbf{\Sigma}_{c}^{T} \mathbf{\Sigma}_{a}^{-1} \mathbf{\Sigma}_{c}$ is the Schur complement of matrix $\mathbf{\Sigma} = \mathbf{\Phi} + \mathbf{W} \mathbf{W}^{T}$  rewritten in the form of the block matrix 
$\mathbf{\Sigma}=\begin{pmatrix}
       \mathbf{\Sigma}_{a}  &   \mathbf{\Sigma}_{c}  \\
     \mathbf{\Sigma}_{c}^{T} & \mathbf{\Sigma}_{b}
\end{pmatrix}$.

In summary, Eq. \ref{eq:bfrprior} gives the prior of the points in latent space $\mathbf{Z}$ with dimension $L$. Eq. \ref{eq:bfract} gives the conditional probability of sampling an actor's facial expression $y_{act}$ with dimension $D \gg L$ given a point $z$ of the latent space $\mathbf{Z}$. Eq. \ref{eq:bfrobs} gives the conditional probability of sampling an observer's facial expression $y_{obs}$ with dimension $D \gg L$ given a point $z$ of the latent space $\mathbf{Z}$.
Finally, Eq. \ref{eq:trans2} returns the conditional probability of sampling a facial expression of the observer given a similar facial expression of the actor and to do that it makes use of both $\mathbf{W}_{act}$ and $\mathbf{W}_{obs}$ in a multivariate normal distribution as previously suggested.

For what concerns the first-order phenomenological latent space $\mathbf{X}_{obs}$, the main issue here is to conceive a generative probabilistic latent space model in which: i) either forward / inverse mapping is allowed; ii) the forward step is a nonlinear and continuous mapping in order to smoothly generate the variety of expressions of the observer. Namely, in probabilistic terms it is necessary to provide a mapping $\mathbf{y}_{obs} = g(\mathbf{x}_{obs}; \mathbf{B}) + \epsilon$, where $\epsilon$ is a zero-mean, isotropic, white Gaussian noise  model and $g(\cdot; \mathbf{B})$ is a continuous nonlinear function. To handle non linearity, the latter can be written as a linear combination of basis functions, $g(\mathbf{x}_{obs}) = \sum_j  \mathbf{b}_j \phi_j(\mathbf{x}_{obs})$. 

The problem of deriving a form for the pdf $P(\mathbf{Y}_{obs} \mid \mathbf{X}_{obs})$ in Eq. \ref{eq:gen} can be formalized  as 
\begin{equation}
\mathbf{\check{y}}_{obs} \sim P(\mathbf{Y}_{obs} \mid \mathbf{g}(\mathbf{X}_{obs}= \mathbf{x}_{obs}; \mathbf{B})).
\label{eq:gen_}
\end{equation}

In this perspective, the mapping parameters of the latent space $\mathbf{B}$ can be learned via  the marginalization 
$P(\mathbf{Y}_{obs} \mid \mathbf{X}_{obs}) = \int P(\mathbf{Y}_{obs} \mid \mathbf{g})  P(\mathbf{g} \mid \mathbf{X}_{obs}) d\mathbf{g}$. This problem  has been solved by Lawrence in terms of the Gaussian Process Latent Variable model (GPLVM, \cite{lawrence2004gaussian}), which can be expressed as a Gaussian density over the observations $\mathbf{Y}_{obs}$, namely a product of Gaussian Processes (one for each of the $D$ data dimensions). Formally, $P(\mathbf{Y}_{obs} \mid \mathbf{X}_{obs}) = \prod_{d}^D \mathcal{N}( \mathbf{y}_d; \mathbf{0}, \mathbf{K})$, where $\mathbf{K}$  is a covariance matrix (or kernel) which depends on the $q$-dimensional latent variables (cf. \cite{lawrence2004gaussian} for a derivation). As a result, an efficient closed form  for the marginal likelihood can be derived \cite{lawrence2004gaussian}:
\begin{equation}
P(\mathbf{Y}_{obs} \mid \mathbf{X}_{obs}) = \frac{1}{\sqrt{(2\pi})^{ND}|\mathbf{K}^{D}|} \exp(-\frac{1}{2} tr (\mathbf{K}^{-1} \mathbf{Y}_{obs} \mathbf{Y}_{obs}^{T})),
\label{eq:gen2}
\end{equation}
\noindent where $\mathbf{Y}_{obs} = \left[ \mathbf{y}_{1,obs}, \cdots , \mathbf{y}_{N,obs} \right]$ is the set of training observations and the elements of the kernel matrix $\mathbf{K}$  are defined by a kernel function $(\mathbf{K})_{i,j} = \phi(\mathbf{x}_{i,obs}, \mathbf{x}_{j,obs})$ (in the simulation a Radial Basis Function kernel, RBF, see Sec. \ref{Sec:simul}).

Once the latent variable model has been learned it is straightforward to obtain the inverse pdf $P(\mathbf{X}_{obs} \mid \mathbf{Y}_{obs})$ in Eq. \ref{eq:code} either through GPLVM standard inversion \cite{lawrence2004gaussian}, or by Monte Carlo sampling approximation from Eq. \ref{eq:gen2}.

Eventually, the matching process summarized by function $\mathcal{D}$ can be conceived as an optimized search in the observer's latent space for determining the optimal state $\mathbf{X}_{obs} = \mathbf{x}^\star_{obs}$ that maximises the similarity between the observed facial expression and one currently generated by using Eq. \ref{eq:gen2} (further details in Sec. \ref{Sec:simul}).

\section{Putting Theory into Work: Results}
\label{Sec:simul}

\subsection{Implementation Details}

\begin{figure}[tb]
	\begin{centering}
	\includegraphics[width=1\textwidth]{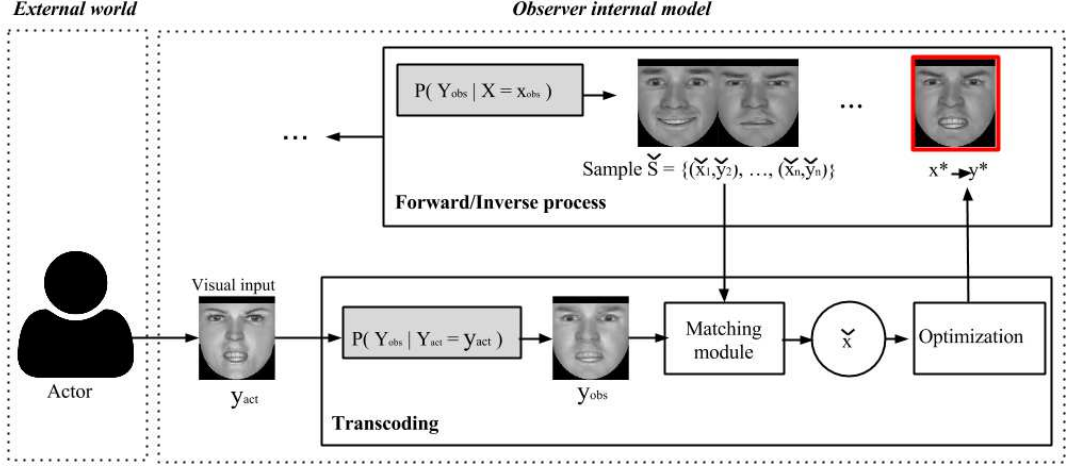}
	\caption{The schema of the proposed computational architecture}
	\label{fig:architecture_schema}
	\end{centering}
\end{figure}

The overall simulation scheme is outlined in Fig. \ref{fig:architecture_schema}.  The visual input $\mathbf{y}_{act}$, namely the observed facial expression of the actor, is transcoded via the self-projection latent space $\mathbf{Z}$ in the observer's internal representation $\mathbf{y}_{obs}$. This mapping is suitably learned so that $\mathbf{y}_{obs}$ will exhibit the same facial expression displayed in $\mathbf{y}_{act}$ but with the identity appearance  of the observer.  At the same time a set $\mathbf{\check{S}}$ of samples  of observer's  expressions is generated through Eq. \ref{eq:gen2}. Along the matching process $\mathcal{D}$,  a similarity measure is used in order to evaluate the likelihood between the samples in $\mathbf{\check{S}}$ and the projected observation $\mathbf{y}_{obs}$. The initial state $\check{\mathbf{x}}$ is selected as: 

\begin{equation}
\check{\mathbf{x}} \in \check{\mathbf{S}} \mid \check{\mathbf{x}} \mapsto \check{\mathbf{y}}_{obs} \land \check{\mathbf{y}}_{obs} = \argmax{y}\mathcal{D}(\check{\mathbf{y}}_{obs},\mathbf{y})
\label{eq:starting_condition}
\end{equation}

Such choice is refined by resorting to a search optimization process in the first-order phenomenological latent space leading to an optimal internal representation $\mathbf{x}^{\star}_{obs}$ of the observed actor's expression $\mathbf{y}_{act}$. 

In the following we provide some simulation details and preliminary results of the proposed system.

\subsubsection{The Transcoding Process}
The aim of this mapping is to generate an internal expression $\mathbf{y}_{obs}$ that exhibits the same facial expression displayed in $\mathbf{y}_{act}$ but with the identity  of the observer, as previously proposed with Eq. \ref{eq:trans2}. In order to simplify the process of parameter learning, it is possible to note that Eq. \ref{eq:bfrobs} basically creates subspaces of $\mathbf{Z}$ for each facial expression of the observer, and that Eq. \ref{eq:bfract} does the same for the facial expression of the actor. Since both Eq. \ref{eq:bfract} and Eq. \ref{eq:bfrobs} share the same latent space $\mathbf{Z}$ and the same set of facial expressions, it is likely that similar facial expression of the observer and the actor would lay on close points in the latent space $\mathbf{Z}$. Thus, the process of parameter learning can be simplified using a process of Principal Component Analysis (PCA) over a training set of the observer performing different expressions and using the estimated mapping parameters in order to project the facial expression of the actor, obtaining a new observation exhibiting the observer identity, but maintaining the actor facial expression.

This procedure is somehow dual to that recently proposed by Mohammadzade and Hatzinakos \cite{Mohammadzade2013Projection} (to which we refer for more in deep technical details), where it was shown that images of different subjects' faces with the same facial expression are located in a common subspace; here, instead the same facial expression is maintained, changing the identity of the subject into a desired one.

Note that for this process it is necessary a \emph{training set} of different facial expressions of the same subject with the desired identity.

The projection leads to a synthesis error, namely the distance between the desired identity and the observed one. The error estimation process requires a validation set. The validation set $\mathbf{V = \{v_j\}_{j=1}^n}$ is the set of $n$ images of different subjects with different identities $\mathbf{i_1, \dots , i_m}$ and displaying the same set of facial expressions used in the training set. Thus, the size of a validation set is $m \times w$, with $m$ the number of validation subjects (i.e. number of identities) and $w$ the number of considered facial expressions (i.e. the cardinality of the training set). 

For each image in $\mathbf{V}$ is associated a ground truth, namely the image in the training set exhibiting the same facial expression of the validation image. We call this set of ground truth images $\mathbf{G}$. 
Thus, for each image in $\mathbf{V}$ it is possible to compute the relative set of synthesis error $\mathbf{\Xi}$ as the differences between the corresponding projected image in $\mathbf{Y}_{obs}$ and the corresponding ground truth in $\mathbf{G}$.

In order to improve the performance of projecting a very similar facial expression and reduce the number of the necessary training images, the face is split in semantic/spatial parts (i.e. the eyebrows, the eyes, the nose, the mouth and the cheeks), thus estimating several transcoding modules for each of these face parts.

\begin{figure}[tb]
	\begin{centering}
	\includegraphics[width=0.65\textwidth]{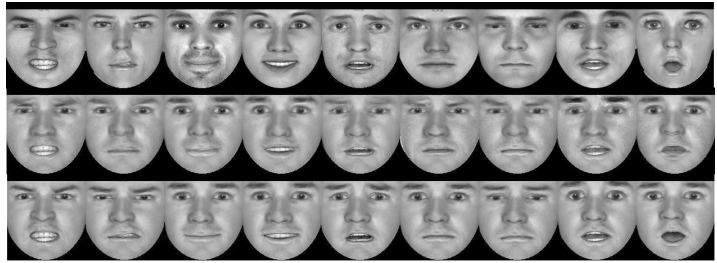}
	\caption{Some examples of synthetic faces projected using self-projection. In the top row the observations, in the middle row the projected faces, whereas in the bottom row the ground truth images.}
	\label{fig:examples_synth}
	\end{centering}
\end{figure}

\begin{figure}[tb]
	\begin{centering}
	\includegraphics[width=0.65\textwidth]{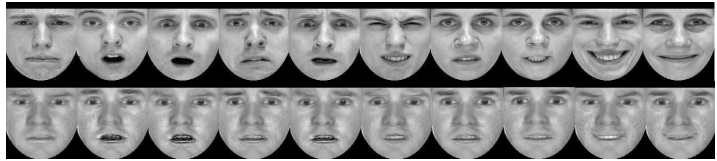}
	\caption{Some examples of real faces projected using self-projection. In the top row the observations, whereas in the bottom row the projected image.}
	\label{fig:examples_real}
	\end{centering}
\end{figure}

Figure \ref{fig:examples_synth} shows examples of projected synthetic images, whereas Fig. \ref{fig:examples_real} shows examples of projected real images  (from the MMI­-Facial Expression Database collected by Valstar and Pantic \cite{valstar2010induced,pantic2005web}). Although we included only $10$ synthetic facial expressions in the training and validation set, the preliminary results are promising; adding more facial configurations might allow a higher degree of generalization and better results.

\subsubsection{The Forward / Inverse Process}

The GPLVM introduced in Sec. \ref{Sec:model} has the capability of learning with few samples and of generating smooth dynamics between points of the latent space (if an appropriate kernel is used, in this case an RBF). For simulation purposes, the GPLVM model  was implemented in the form of a Hierarchical Gaussian Process Latent Variable Model (HGP-LVM) \cite{lawrence2007hierarchical}. The HGP-LVM is an extension of the original Gaussian Process Latent Variable Model (GP-LVM) \cite{lawrence2004gaussian}.  We choose to use this tool to introduce hierarchical constraints based on the different parts composing the face (i.e. eyebrows, eyes, nose, mouth and cheeks). For technical details about this technique, see \cite{lawrence2007hierarchical}.

Using  this model it is possible to sample a vocabulary $\mathbf{S: X \mapsto \{{Y}_{obs}^j\}_{j=1}^\infty}$ that generates an infinite set of synthetically generated facial expressions of a specific desired identity from $q$-dimensional internal representations. A subsample $\mathbf{\check{S}}$ of such vocabulary is used by the matching module (Fig. \ref{fig:architecture_schema}).

Each point of the  latent space represents an internal phenomenological state $\mathbf{x}$. Such representation has dimension $q$ (in our case $2$-dimensional). The likelihood between the projected observation $\mathbf{y}_{obs}$ and the observations generated from sampled latent points $\mathbf{\check{x}}$ provide to us an approximation of the conditional distribution $\mathbf{P( X \mid Y_{obs} = \mathbf{y}_{obs} )}$, that represents the activation levels of the topology of such first-order phenomenological latent space. 

\begin{figure}[tb]
	\begin{centering}
	\includegraphics[width=0.5\textwidth]{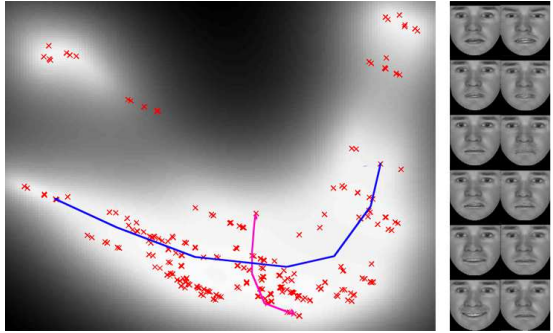}
	\caption{Example of a latent space using the HGP-LVM. On the right examples of 12 sampled generated images from the trajectories highlighted on the shown latent space.}
	\label{fig:hgplvm_space}
	\end{centering}
\end{figure}

In Fig. \ref{fig:hgplvm_space} is shown a latent space generated from a high number of different facial expressions ($238$) and examples of generated facial expressions from points of such latent space.

\subsection{Matching Module and Optimization Process}
\label{sec:SSIM}

Since for simulation purpose the expression representation were in the form of images, we used as similarity index the Structural Similarity (SSIM) measure \cite{wang2004image}. On the basis of the statistical model behind the mapping,  one can conceive the projected images  as noisy representations of the ground truth images; in this perspective, the SSIM  was shown to be a  consistent measure \cite{wang2004image}. Additional tests with different similarity measures (e.g. Pearson's correlation measure) produced poorer results, further motivating this choice. In a perspective of future real time applications, measures with a high computational cost were not considered.

The SSIM metric is processed among different windows of an image and then the average among them is used as a final measure. The measure between two windows $x$ and $y$ of size $N \times N$ (in our model a Gaussian window of $80 \times 80$ pixels with sigma $3$) is given by:

\begin{equation}
SSIM(x,y) = \frac{(2\mu_x\mu_y + c_1)(2\sigma_{xy} + c_2)}{(\mu_2^x + \mu_y^2 + c_1)(\sigma_x^2 + \sigma_y^2 + c_2)}
\label{eq:SSIM}
\end{equation}

Where $\mu_x$ and $\mu_y$ are respectively the means of $x$ and $y$, $\sigma_x^2$, $\sigma_y^2$ and $\sigma_{xy}$ are respectively the variances of $x$ and $y$ and the covariance of $x$ and $y$, $c_1$ and $c_2$ are two variables to stabilize the division with weak denominator (see \cite{wang2004image} for more technical details).

As it is possible to sample only a finite number of points of the latent space (and corresponding synthetically generated images), the most likely point with respect to the projected image $\mathbf{y}_{obs}$ (Eq. \ref{eq:starting_condition}) was used as the initial condition of an interior--point method optimization \cite{nesterov1994interior}. This optimization method refines the selected position $\check{\mathbf{x}}$ within the latent space in order to find the closer position $\mathbf{x}^\star$ that generates the image with maximum value of SSIM respect to $\mathbf{y}_{obs}$.

\subsection{Dataset}

The dataset has been generated by using the FaceGen software\footnote{http://www.facegen.com/}. Synthetically generated images were used in order to simplify the problem, as with the current datasets of facial expressions it was not possible to generate a sound validation set; as a matter of fact the different subjects in the datasets do not display the very same facial configuration, making it impossible to find conclusive correspondences between observations.

\begin{figure}[tb]
	\begin{centering}
	\includegraphics[width=0.8\textwidth]{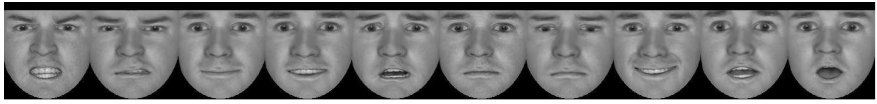}
	\caption{The training images used for the tests of our architecture.}
	\label{fig:gt}
	\end{centering}
\end{figure}

The dimension of the images used was  $140 \times 154$ pixels. For the sake of simplicity,  in these preliminary tests  only $10$ facial expressions were considered, namely anger, annoyance, delight, fake smile, fear, neutral, sadness, smile, surprise and wonder (Fig. \ref{fig:gt}). Note that these labels were subjectively attributed, but what is important for our simulation is to have different perceptual stimuli, as at this stage the intention is to provide a model for their internal representation and not for affect recognition. These facial expressions were selected as the most representative with respect to the ones obtainable using the FaceGen software. We selected the whole list of emotional expressions available in the software plus further combinations of them. Some additional features were added to the basic emotional expressions when necessary; for example, a 50\% look down expression in sadness. Most important, this list of expressions included very perceptually similar stimuli, such as the fake smile and the smile, neutral and sadness, and surprise and wonder.

The dataset includes:
\begin{itemize}
\item A training set of $10$ images, namely a subject\footnote{Note that in order to be consistent with the studies on ST and MNS, the subject used in the training set has to be the same for all the individuals. A possible and more suitable ``shared'' subject should be the average identity obtainable by averaging different identities with the same facial expression. However, in this work the problem is simplified by using one specific identity.} exhibiting all the selected facial expressions (Fig. \ref{fig:gt});
\item A validation/test set of $28$ subjects, each one exhibiting the $10$ selected facial expressions (some examples in the top row of Fig. \ref{fig:examples_synth});
\end{itemize}

A 4-fold cross validation approach was used, thus for each test we split the validation/test set in 2 sets: a validation set of $21$ subjects ($21 \times 10 = 210$ validation images) and a test set of $7$ subjects ($7 \times 10 = 70$ test images).

The training set was used by the transcoding module and by the forward/inverse process to estimate mapping parameters. The validation sets were exploited to estimate the synthesis error of the transcoding module. The test sets were used to obtain the results provided in the next sections.

\subsection{Evaluation}

\subsubsection{Quality of Reconstruction}

Using the measure introduced in Sec. \ref{sec:SSIM}, the evaluation process requires to determine the structural similarity between the projected images $y_{obs}$ and the relative ground truth images $g \in G$, and the structural similarity between the images $\check{y}$ generated from their internal representations $x^\star$ and their corresponding ground truth images $g \in G$. In order to have a baseline for comparisons, we also evaluated the structural similarity between the observations $y_{act}$ and the associated ground truth images $g \in G$.

\begin{figure}[ht]
	\begin{centering}
	\includegraphics[width=0.6\textwidth]{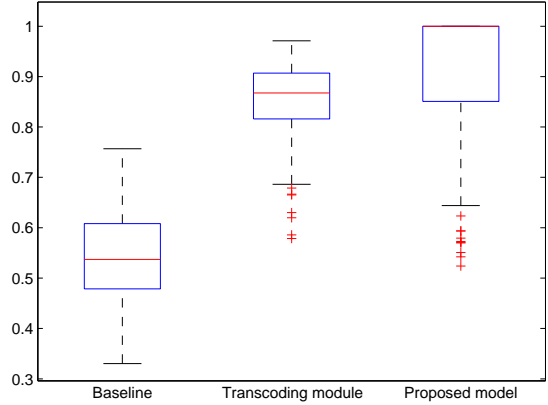}
	\caption{Quality of Reconstruction Boxplot.}
	\label{fig:qor}
	\end{centering}
\end{figure}

The execution of the 4 cross validation tests produced the results summarized in Table \ref{tab:qor_results} and in Fig. \ref{fig:qor}.

\begin{table}[ht]
	\begin{centering}
		\begin{tabular}{lccc}
		\hline
		Case            & Mean & Median  & Std \\
		\hline
		Baseline      & $0.5425$ & $0.5370$ & $0.0871$ \\
		Transcoding module only  & $0.8525$ & $0.8674$ & $0.0713$ \\
		Proposed model & $0.9148$ & $0.9999$ & $0.1374$ \\
		\hline
		\end{tabular}
		\caption{Quality of Reconstruction results (likelihoods).}
		\label{tab:qor_results}
	\end{centering}
\end{table}

As the maximum similarity for the selected measure (SSIM) can be $1$ if and only if the two images under evaluation are identical, we define a \emph{reconstruction error} as the difference between $1$ and the measure of similarity. Thus, the average reconstruction error was reduced from $0.46$ to $0.15$ by using the transcoding module and to $0.09$ by using the whole model. This means that the proposed model decreased the reconstruction error by approximately 68\% using the transcoding module only and by approximately 81\% by exploiting the whole proposed model.

Furthermore, the proposed model provide a median of $0.9999$ of SSIM with respect to the relative ground truth images. This impressive result is due to the fact that the learned first-order phenomenological latent space exhibits an impressive ability to reconstruct the training images ($\check{y} \approx g$) and structural dissimilarities are mainly due to some noisy transcoded images $y_{obs}$, that in turn leads to non optimal selection of $x^\star$ and its corresponding non optimal image $\check{y}$ (this can be noted also by the higher standard deviation).

\subsubsection{Classification}

As  explained from the very beginning, the focus of this work is \emph{not} the classification of emotions. However, we understand a long term application of the model will be that of classifying facial expressions observations in terms of affective labels. Thus, we also measured the classification performance with respect to the $10$ possible facial expressions in our dataset.

Given a set $G = \{g_j\}_{j=1}^{10}$ of ground truth images for each class $j$, these have a corresponding set of latent positions $\mathbf{x_G^j}$ in the latent space. Given a latent point $\mathbf{x}^\star$ estimated from a new observation $\mathbf{y}_{act}$, the class $c$ of $\mathbf{x}^\star$ (and so of $\mathbf{y}_{act}$) is $j$ of $\mathbf{x_G^{j\star}}$ spatially closer to $\mathbf{x}^\star$.

For each test of our $4$-fold cross validation process we computed a confusion matrix relative to the classification of the $70$ test images. The four confusion matrixes were summed, obtaining a new matrix relative to the $280$ images used for validation and test of our architecture.

In order to estimate the baseline confusion matrix we classified each observation $\mathbf{y}_{act}$ in the validation/test set with respect to their structural similarity with the ground truth images in $\mathbf{G}$. So, in a baseline perspective, the class of the observation $\mathbf{y}_{act}$ is the class of the ground truth image in $\mathbf{G}$ that is more structurally similar to $\mathbf{y}_{act}$.

In Table \ref{tab:conf_matr_baseline} and \ref{tab:conf_matr_full_arc} are respectively the confusion matrix of the baseline and the one obtained using the proposed architecture.

\begin{table}[p]
	\begin{centering}
	\scalebox{0.75}{
		\begin{tabular}{l|cccccccccc}
		\hline
		Expression            & Anger & Annoyance  & Delight & Fake Smile & Fear & Neutral & Sadness & Smile & Surprise & Wonder \\
		\hline
		Anger      & $20$ & $1$ & $3$ & $4$ & $0$ & $0$ & $0$ & $0$ & $0$ & $0$  \\
		Annoyance  & $0$ & $11$ & $7$ & $2$ & $1$ & $2$ & $0$ & $0$ & $3$ & $2$ \\
		Delight & $0$ & $0$ & $25$ & $1$ & $0$ & $0$ & $0$ & $0$ & $2$ & $0$ \\
		Fake Smile & $1$ & $0$ & $3$ & $18$ & $0$ & $0$ & $0$ & $6$ & $0$ & $0$ \\
		Fear & $0$ & $0$ & $4$ & $0$ & $16$ & $3$ & $0$ & $0$ & $4$ & $1$ \\
		Neutral & $0$ & $0$ & $15$ & $0$ & $0$ & $9$ & $2$ & $0$ & $2$ & $0$ \\
		Sadness & $0$ & $0$ & $14$ & $0$ & $0$ & $11$ & $2$ & $0$ & $1$ & $0$ \\
		Smile & $1$ & $0$ & $6$ & $9$ & $0$ & $0$ & $0$ & $12$ & $0$ & $0$ \\
		Surprise & $0$ & $0$ & $5$ & $0$ & $1$ & $5$ & $0$ & $0$ & $17$ & $0$ \\
		Wonder & $0$ & $0$ & $0$ & $0$ & $1$ & $6$ & $0$ & $0$ & $4$ & $17$ \\
		\hline
		\end{tabular}
	}
		\caption{Confusion matrix using the baseline.}
		\label{tab:conf_matr_baseline}
	\end{centering}
\end{table}

\begin{table}[p]
	\begin{centering}
	\scalebox{0.75}{
		\begin{tabular}{l|cccccccccc}
		\hline
		Expression            & Anger & Annoyance  & Delight & Fake Smile & Fear & Neutral & Sadness & Smile & Surprise & Wonder \\
		\hline
		Anger      & $23$ & $1$ & $1$ & $2$ & $0$ & $0$ & $1$ & $0$ & $0$ & $0$  \\
		Annoyance  & $0$ & $24$ & $0$ & $0$ & $0$ & $2$ & $1$ & $0$ & $0$ & $1$ \\
		Delight & $0$ & $0$ & $20$ & $1$ & $0$ & $5$ & $0$ & $0$ & $2$ & $0$ \\
		Fake Smile & $0$ & $0$ & $0$ & $22$ & $0$ & $0$ & $0$ & $6$ & $0$ & $0$ \\
		Fear & $0$ & $1$ & $1$ & $0$ & $17$ & $5$ & $0$ & $0$ & $3$ & $1$ \\
		Neutral & $0$ & $0$ & $1$ & $0$ & $1$ & $24$ & $2$ & $0$ & $0$ & $0$ \\
		Sadness & $0$ & $2$ & $2$ & $0$ & $0$ & $17$ & $7$ & $0$ & $0$ & $0$ \\
		Smile & $0$ & $0$ & $1$ & $8$ & $0$ & $0$ & $0$ & $19$ & $0$ & $0$ \\
		Surprise & $0$ & $1$ & $1$ & $0$ & $2$ & $10$ & $0$ & $0$ & $13$ & $1$ \\
		Wonder & $0$ & $1$ & $0$ & $0$ & $2$ & $5$ & $0$ & $0$ & $1$ & $19$ \\
		\hline
		\end{tabular}
	}
		\caption{Confusion matrix using the whole proposed model.}
		\label{tab:conf_matr_full_arc}
	\end{centering}
\end{table}

\afterpage{\clearpage}

\begin{figure}[p]
\begin{subfigure}{.5\textwidth}
  \centering
  \includegraphics[width=0.9\linewidth]{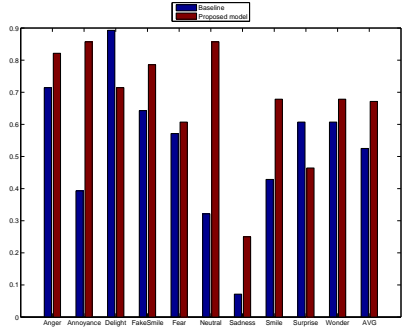}
  \caption{Sensitivity}
  \label{fig:sensitivity}
\end{subfigure}
\begin{subfigure}{.49\textwidth}
  \centering
  \includegraphics[width=0.9\linewidth]{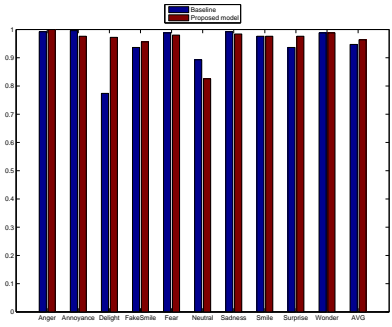}
  \caption{Specificity}
  \label{fig:specificity}
\end{subfigure}
\begin{subfigure}{.5\textwidth}
  \centering
  \includegraphics[width=0.9\linewidth]{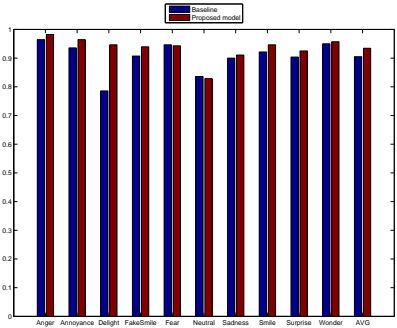}
  \caption{Accuracy}
  \label{fig:accuracy}
\end{subfigure}
\begin{subfigure}{.49\textwidth}
  \centering
  \includegraphics[width=0.9\linewidth]{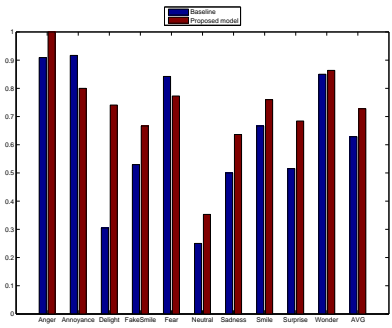}
  \caption{Precision}
  \label{fig:precision}
\end{subfigure}
\begin{subfigure}{1\textwidth}
  \centering
  \includegraphics[width=0.5\linewidth]{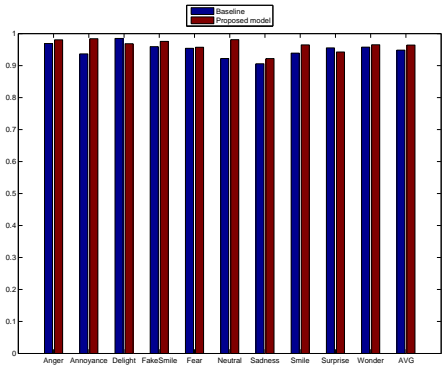}
  \caption{Negative Predictive Value}
  \label{fig:npv}
\end{subfigure}
\caption{Plots of the classification measures for the proposed model.}
\label{fig:classification_measures}
\end{figure}

We computed the sensitivity, specificity, accuracy, precision and negative predictive value (NPV) of each class for both the baseline method and the proposed model. The results are summarized in Table \ref{tab:class_results} (averages of the $10$ classes) and illustrated in Fig. \ref{fig:classification_measures}.

\begin{table}[p]
	\begin{centering}
	\scalebox{0.8}{
		\begin{tabular}{lccccc}
		\hline
		Case            & Sensitivity & Specificity  & Accuracy &  Precision & NPV\\
		\hline
		Baseline      & $0.5250$ &  $0.9472$ & $ 0.9050$ & $0.6283$ & $0.9483$ \\
		Proposed model & $0.6714$ & $0.9635$ & $0.9343$ & $0.7277$ & $0.9641$ \\
		\hline
		\end{tabular}
	}
		\caption{Results for the enhancement in classification (averages).}
		\label{tab:class_results}
	\end{centering}
\end{table}

The proposed model overcomes the baseline approach with respect to all the considered measures (averages). In order to verify the significance of such differences (i.e. for each measure all the single performance of each class with the baseline and with the proposed model), we tested them with a paired Wilcoxon signed-rank test \cite{wilcoxon1945individual}, as it was not possible to assume a normally distributed population.

The increment of sensitivity and NPV results are statistically marginally significant (p-values = $0.0537$ and $0.0674$), whereas the increment in accuracy and precision results statistically highly significant (p-values = $0.0049$ and $0.0322$). The increment in specificity results statistically not significant (p-value = $0.7646$). These results motivate us to investigate further possible applications of such model in the affect recognition domain; however, we are conscious of the fact that these results are currently just preliminary and far from giving a robust proof of such ability, that, as we mentioned from the very beginning, is not the final target of this study.

\section{Conclusion and Future Works}
\label{Sec:concl}

In this work we investigated current Simulation Theories and we provided a theoretical and computational account. We limited our investigation to the sole process of mapping an external facial expression of a target to its corresponding internal state of an observer. Developing a computation model for this process allows us to  build  novel methods for social agents and robots to understand people's facial expressions.

The proposed model might be used to solve some of the current challenges of affect recognition systems. In fact, for the computational simulation of such model we did not use any \emph{a priori} affective label. We based our learning process on the appearance of facial expressions, leaving the attribution of the affective state to a further step. This final step may be attained by estimating the \emph{as-if} simulated mental state, given the internal representation of the observed simulus and contextual factors. The estimation might be realized through direct social experiences of the observer, giving to the attribution of mental states the cultural and contextual connotations that lacks in current affect recognition systems.

This work shown that a simulation process can be achieved by the use of two modules: one mapping the external perceived facial expression to an internal representation of such stimulus, and one mapping this transcoded stimulus to a first-order phenomenological latent space.

Significantly, the proposed model can be extended to other different kind of overt behaviours. Gestures, posture and vocal cues can be simulated with the same proposed process and mapped to other self-projection and first-order phenomenological latent spaces. In fact, these signals can be treated as vectors, similarly to the used facial expression images. Then, it is likely that a self-projection latent space of a speech signal would map the pitch of the voice of the actor to that of the observer, maintaining the same principal frequencies. A similar outcome would be realized for gesture and posture signals, where physical constraints of the actor's body would be mapped to those of the observer's body. These self-projection latent spaces can be associated to very similar first-order phenomenological latent spaces. Such internal representations, shared among different modalities and easily treatable from a probabilistic point of view, might solve the challenging problem of multimodality integration in affect systems \cite{zeng2009survey}.

Another important feature of such model is that of generalizing over similar observed stimuli. In fact, due to intrinsic features of GPLVM, similar observations lie on close points of the latent space. Thus, instead of taking just one internal representation $x^\star$ it might be taken the whole pdf given by Eq. \ref{eq:code}. This gives the opportunity to generalize over similar perceptual stimuli so to reduce the overall error.

Finally, the proposed model can be easily extended to consider temporal dynamics. Such extension might make use of Markov Chains or Dynamic Bayesian Networks.

Some limitations of this work are mostly due to experimental choices at the algorithmic level. First, the considered facial expressions are well aligned frontal faces. This is not a problem per-se, but we understand that a future work will be to generalize over different alignments of such faces. Second, the used stimuli are synthetic facial expressions. Thus, we were not able to provide quantitative results for real facial expressions. However, in Fig. \ref{fig:examples_real} we presented some first qualitative examples of transcoded real facial expressions. Considering the use of just $10$ different facial expressions for the estimation of the model's parameters these are promising results, and might allow affect recognition researcher to use synthetic facial expression datasets instead of costly and hard to create real images datasets.

Our current efforts are directed to the problem of generalizing the model to other modalities, while addressing the step of mental state attribution.

\label{sect:bib}
\bibliographystyle{plain}
\bibliography{bibliography}

%------------------------------------------------------------------------------

\end{document}